\documentclass[letterpaper]{article} 
\usepackage{aaai24}  
\usepackage{times}  
\usepackage{helvet}  
\usepackage{courier}  
\usepackage[hyphens]{url}  
\usepackage{graphicx} 
\usepackage{natbib}  
\usepackage{caption} 
\usepackage{algorithm}
\usepackage{algorithmic}
\usepackage{booktabs}
\usepackage{amssymb}

\usepackage{newfloat}
\usepackage{listings}
\DeclareCaptionStyle{ruled}{labelfont=normalfont,labelsep=colon,strut=off} 
\floatstyle{ruled}
\newfloat{listing}{tb}{lst}{}
\floatname{listing}{Listing}
\title{Hierarchical Aligned Multimodal Learning for NER on Tweet Posts}
\author{
    Peipei Liu\textsuperscript{\rm 1,\rm 2},
    Hong Li\textsuperscript{\rm 1,\rm 2}\thanks{The corresponding author},
    Yimo Ren\textsuperscript{\rm 1,\rm 2},
    Jie Liu\textsuperscript{\rm 1,\rm 2}, 
    Shuaizong Si\textsuperscript{\rm 1}, 
    Hongsong Zhu\textsuperscript{\rm 1,\rm 2}, 
    Limin Sun\textsuperscript{\rm 1,\rm 2}
}
\affiliations{
    \textsuperscript{\rm 1}Institute of Information Engineering, Chinese Academy of Sciences\\
    19 Shucun Road, Haidian District, Beijing 100085 P.R.China\\
    \textsuperscript{\rm 2}School of Cyber Security, University of Chinese Academy of Sciences\\

    19 Yuquan Road, Shijingshan District, Beijing 100049 P.R.China\\
    \{liupeipei, lihong, zhuhongsong\}@iie.ac.cn
}

\usepackage{bibentry}
\usepackage{amsmath,graphicx}
\usepackage[dvipsnames]{xcolor}
\usepackage{multirow}
\usepackage{graphicx}

\begin{document}
\maketitle
\begin{abstract}
    Mining structured knowledge from tweets using named entity recognition (NER) can be beneficial for many downstream applications such as recommendation and intention understanding. With tweet posts tending to be multimodal, multimodal named entity recognition (MNER) has attracted more attention. In this paper, we propose a novel approach, which can dynamically align the image and text sequence and achieve the multi-level cross-modal learning to augment textual word representation for MNER improvement. To be specific, our framework can be split into three main stages: the first stage focuses on intra-modality representation learning to derive the implicit global and local knowledge of each modality, the second evaluates the relevance between the text and its accompanying image and integrates different grained visual information based on the relevance, the third enforces semantic refinement via iterative cross-modal interactions and co-attention. We conduct experiments on two open datasets, and the results and detailed analysis demonstrate the advantage of our model.
\end{abstract}

\section{Introduction}

In recent years, social media platforms like Twitter and Facebook have made great development, and they provide important sources (i.e., tweets) for various applications such as the identification of cyber-attacks or natural disasters, analyzing public opinion, and mining disease outbreaks \cite{Bruns2012ToolsAM, Ritter}. Named entity recognition, the task of detecting and classifying named entities from unstructured free-form text, is a crucial step to extract the structured information for those downstream applications \cite{Perera2018CyberattackPT, CDTDNN}. With tweets tending to be multimodal and traditional unimodal NER methods having challenges to understand these multimedia contents perfectly, multimodal named entity recognition (MNER) has become a new direction and\begin{figure}[h]
  \scriptsize
    \centering
    \includegraphics[height=1.5in,width=3.28in]{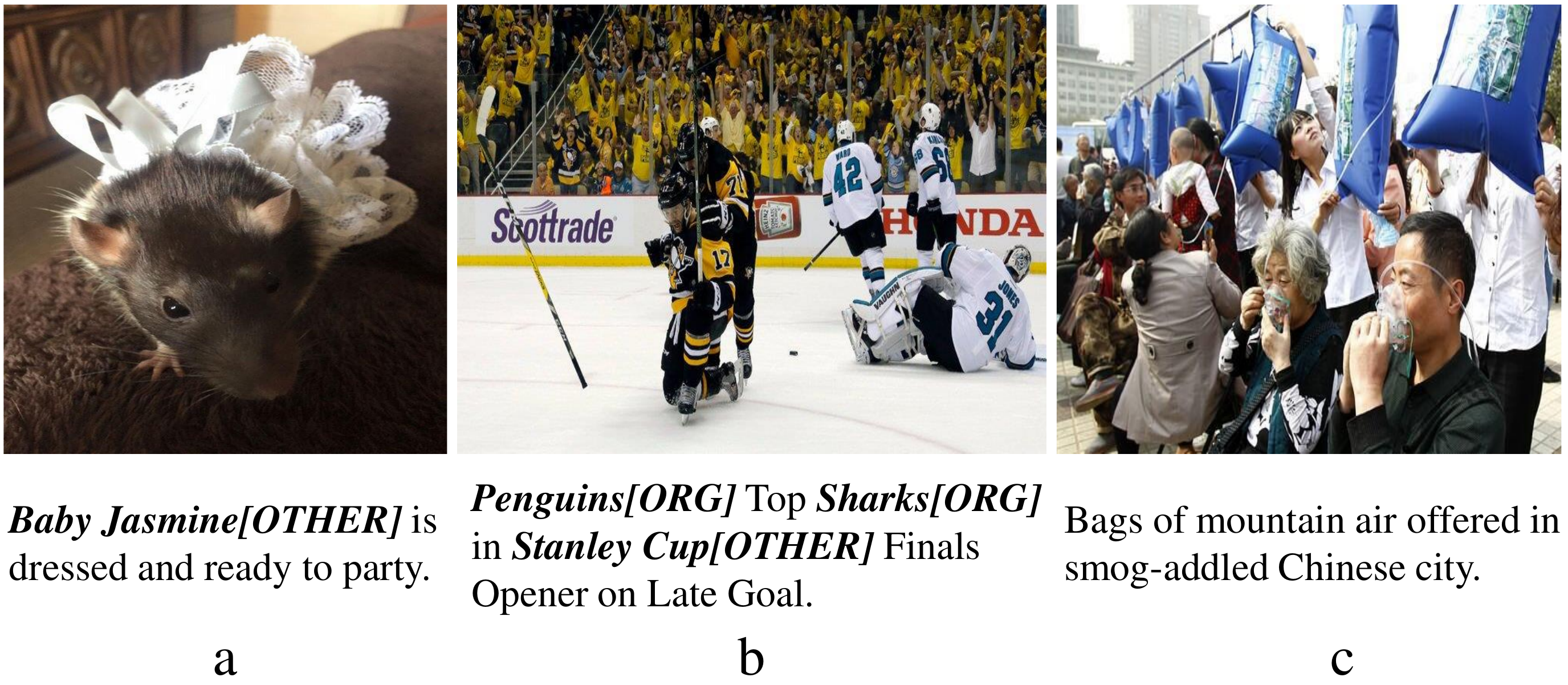}
  \caption{The samples for MNER task, where the named entities and their types are highlighted. a: fully relevant (explicit support information), b: partially relevant (implicit support information), c: entity irrelevant (no entities, no support).}
    \label{introsample}
    \vspace{-0.3cm}
  \end{figure} it improves conventional text-based NER by considering images as additional inputs \cite{lu2018visual, yu2020improving, sun2021rpbert, Chen2022HybridTW}. For example, as shown in Figure \ref{introsample}(a), the supplement of visual information can alleviate the semantic ambiguity and inadequacy problem caused by only text information while classifying the named entity \textit{Baby Jasmine} to \textbf{OTHER} instead of \textbf{PER}.

The core of existing MNER methods is to achieve the fusion and alignment of visual information and textual information through different cross-modal technologies. The methods can roughly be divided into four main streams: (1) \cite{lu2018visual, zhang2018adaptive, omer2019aitr, Durant2021MGC, catmner2022} employ pre-trained CNN models such as ResNet \cite{he_resnet} to encode the whole images into a global feature vector, and then augment each word representation with the global image vector by effective attention mechanism. (2) \cite{yu2020improving, xu2022maf, Liu2022MultiGranularityCR} divide the feature map obtained from the whole image into multiple blocks averagely, and subsequently learn the most valuable vision-aware word representation by modeling the interaction between text sequence and the visual regions with Transformer or gating mechanism.  (3) Some researchers apply object detection models like Mask RCNN \cite{mrcnn_he} to obtain the visual objects from the associated image, and then combine the object-level visual information and textual word information based on GNN or cross-modal attention \cite{OCSGA2021MM,Zhang2021UMGF, flatmner2022,chen2022HVPM, Chen2022HybridTW,umgf2}. (4) There are also some works to explore the derivative knowledge of image content including OCR, image caption, query guidance and other image attributes \cite{ASMNER2022, MRC2022MM, ita2022, zhao2022DTIP, PromptMNER,querymultivision23ai}, which is used to guide words to get the expanded visual semantic information.

Despite the impressive results of these existing methods, there are still several evident limitations remained. Firstly, the current methods heavily relied on the argument that the accompanying image of posted text is entity-related and the visual information is helpful for textual entity extraction. However, the viewpoint is not always valid and the relevance between the text and image is in various situations: fully relevant, partially relevant and irrelevant. As a result, the noise of irrelevant visual content would lead to misleading interaction representation and further affect the MNER performance. In fact, we can observe that the image adds no additional content to the text in 33.8\% of tweets from the report of \cite{vempala2019tir}. Secondly, the current methods usually leverage the image or object representation directly extracted from the original vision view but the implicit knowledge such as image scene and interactive relationship between different objects is neglected. Take Figure \ref{introsample}(b) as an example, the visual stadium scene can easily help us make a correct prediction for the entities \textit{Penguins} and \textit{Sharks} with regarding them as \textbf{ORG} entities rather than \textbf{OTHER} entities (i.e., not animals). Lastly, although they have achieved state-of-the-art results with cross-modal interaction in various ways, seldom of them have explored the multi-level semantic alignments between the vision and text modality. In practice, there are two key points that a word is the basic unit and several words make up a sentence while an image is composed of a number of objects and attributes. Constructing different level alignments not only captures fine-to-coarse correlations between images and texts, but also takes the advantage of the complementary information among these semantic levels.

There is some work carried out for the partial limitations: \cite{xu2022maf,sun2021rpbert,sun2020riva,extradata} design additional classification tasks to measure the text-image relationship by introducing external tools and datasets, \cite{Liu2022MultiGranularityCR, chen2022HVPM, querymultivision23ai} initially employ the multi-level vision information. However, their solutions are imperfect since additional tasks usually need to be built and the multi-level textual content is not used. Motivated by such findings, we propose the novel \textbf{H}ierarchical \textbf{a}ligned \textbf{m}ultimodal \textbf{Learning} (\textbf{HamLearning}), which aims to end-to-end model multi-level semantics for both modalities and enforce cross-modal interactions at different semantic levels with the basis of measuring the relevance of text and image. Specifically, we perform the model within three stages: 1) Firstly, we use the textual Transformer \cite{vaswani2017attention} to learn the global representation of sentence and contextual representations of words. At the same time, two separate visual encoders are deployed to capture object-to-object relations and consider the global scene of image from semantic and spacial view, respectively. 2) Secondly, the relevance between text sentence and its accompanying image is measured through the global content alignment. Then, we integrate the object-level and image-level visual representation to acquire local-to-global sufficient visual feature based on the relevance score. 3) Finally, we implement cross-modal interaction between word representations and the fused visual feature iteratively to refine the most effective multimodal clues for decoding.

We conduct the extensive experiments on two popular MNER datasets, Twitter2015 \cite{lu2018visual} and Twitter2017 \cite{zhang2018adaptive}, to evaluate the performance of the \textbf{HamLearning} framework, and results show the superiority of our approach. Moreover, the full experimental analyses help us understand the advantages and details of the model comprehensively.

The main contributions of this paper can be summarized as:
\begin{itemize}
  \item We construct multi-level alignments to capture coarse-to-fine interactions between vision and language, and take the advantage of the complementary information among these semantic levels.
  \item We introduce the new spatial and semantic learning of visual scene for the MNER task and directly utilize learned visual feature without using the existing generation tools for visual semantic.
  \item We design the end-to-end dynamic relevance measuring on image-text for specific MNER task instead of performing additional text-image relationship classification tasks based on the external tools and datasets.
  \item Through detailed experiments and analyses, we demonstrate the competitive performance of our method in comparison with the current excellent models.
  \end{itemize}

\section{Related Work}
As social media posts become more multimodal, MNER is attracting researchers' attention. \cite{yu2020improving} extends the vanilla Transformer to cross-modality Transformer for capturing multimodal interactions between text words and image regions, and further designs an additional entity-span detection module to alleviate the bias of visual factor and improve the performance of MNER. \cite{dasfaa2021,catmner2022} enhance the text representation by integrating the knowledge and attributes of corresponding images. Different from above works of using the whole image, \cite{OCSGA2021MM, Zhang2021UMGF, umgf2, Chen2022HybridTW} believe that the textual entities are determined by the visual objects, so the object-level visual representation is used to guide the entity recognition in the text. In addition to improvements of MNER methods, \cite{sun2020riva,sun2021rpbert} control the effect of images on text at two different stages through gate- and attention- mechanisms, and pre-train multimodal BERT models for MNER based on text-image relationship inference.

\begin{figure*}[h]
  \centering
  \includegraphics[height=3.0in,width=6.6in]{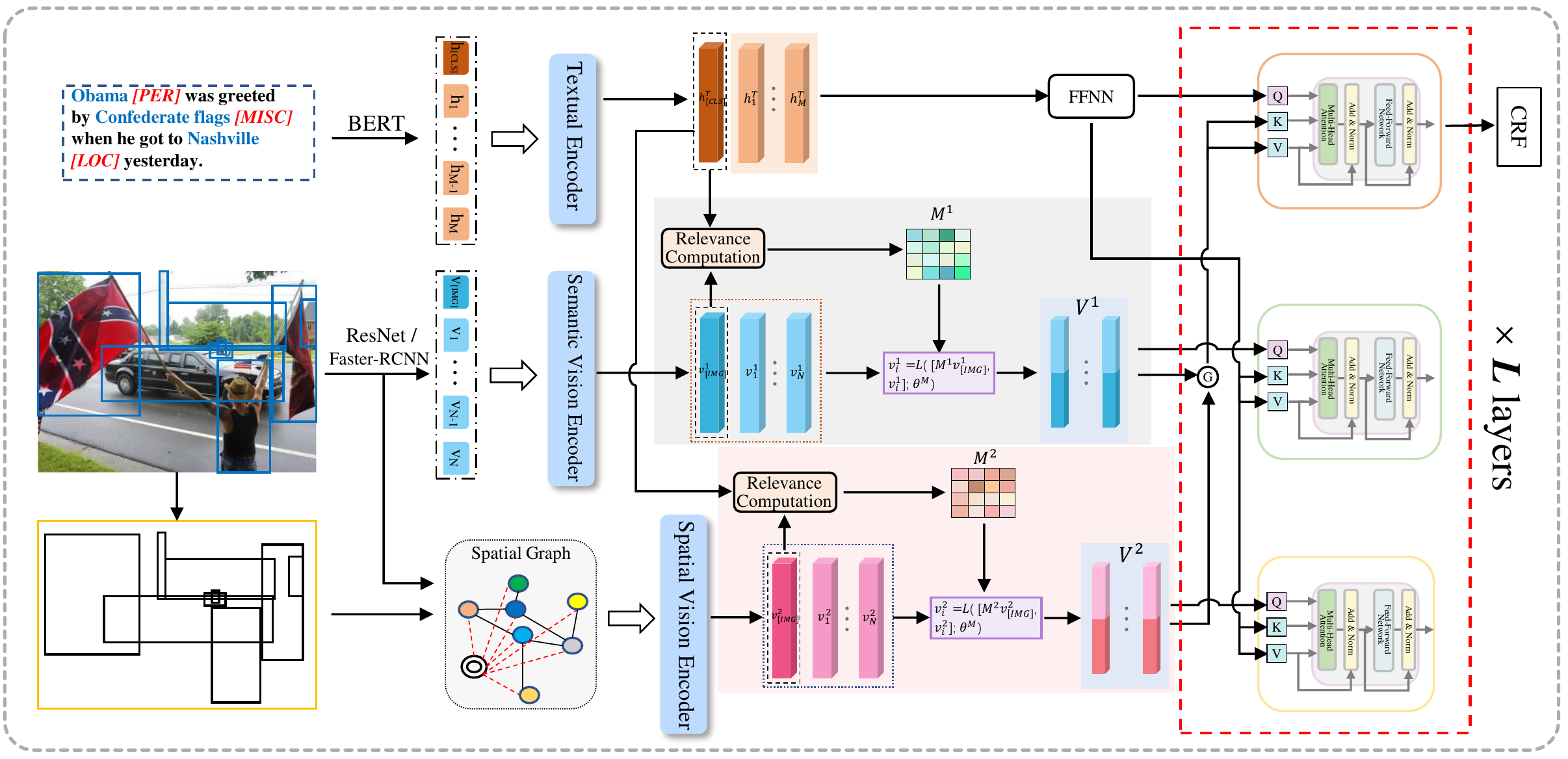}
  \caption{The overview of our proposed method.}  
  \label{model_overview}
  \vspace*{-0.4cm}
\end{figure*}

Nevertheless, these researches just focus on the certain visual grained feature (i.e., fine or coarse), but ignore the effect of multi-level interaction. \cite{chen2022HVPM, Chen2022HybridTW, querymultivision23ai} make a few preliminary attempts on multi-level visual information. In their works, they simply exploit the output features from different stages of pre-trained vision models for cross-modal fusion. There are some other approaches that do not directly use the visual information from the images, but they open the new paths to mine the hidden information behind the image. \cite{PromptMNER} designs several prompt templates for each image to bridge the gap between vision space and text space while \cite{MRC2022MM,querymultivision23ai} construct the queries for the image description. Not only that, \cite{ita2022, zhao2022DTIP} introduce the image caption and OCR information to compensate for the irrationality of original visual information.

Some works \cite{xu2022maf,sun2021rpbert,sun2020riva,extradata} also question the matching problem between images and text information, and they have payed attention to the solution. They mainly use the pre-training Vision-Language Model (VLM) such as CLIP \cite{clip} to design additional text-image relationship classification tasks, and add vision assistance to text information based on prediction probability. However, this line of thinking depends too much on the pre-trained VLM models, and the performance has a great correlation with VLM. In addition, the computed vector seems to refine the visual information related to the entities, but in fact, the content of whole image is the main one. If the image does not match with the text, there is still large visual misleading noise.

In this study, we consider such problems comprehensively and create the \textbf{HamLearning} for the solution.

\section{Method}
In this section, we firstly define the MNER task and then we take the image and text sequence shown in Figure \ref{model_overview} as a running example to introduce details of our proposed approach.

\textbf{Task Definition:} Given the input pair containing a text sentence $\textbf{X}$ and an image $\textbf{I}$, the goal of MNER is to detect a set of entities from $\textbf{X}$, and classify them into the pre-defined types \cite{yu2020improving}. As with other works in the literature \cite{moon2018multimodal,lu2018visual,zhang2018adaptive,omer2019aitr,yu2020improving,MRC2022MM,Chen2022HybridTW}, we regard the MNER as a sequence labeling task. Let $\textbf{X}={\{x_1,...,x_M\}}$ denote the input sequence with $M$ words and $\textbf{Y}={\{y_1,...,y_M\}}$ indicate the corresponding label sequence, where $y_i \in {\zeta}$ and ${\zeta}$ is a pre-defined label set in standard BIO2 formats \cite{tjong1999}.
\subsection{Intra-modality Learning}
In this section, we use the Transformer and modified R-GCN to learn local and global representations of text and vision respectively.

\subsubsection{Text Encoding}
\label{textencoding}
BERT \cite{devlin2019bert} benefits from a large external corpus and has a strong dynamic feature extraction capability for the same word in different contexts. In this work, each text sequence ${\textbf{X}}=\{x_1,...,x_M\}$ is fed into the pre-trained 12-layers BERT to get the sequence representations. As we all know, the additional special token “[CLS]” should be added to the first position to represent the global semantic of entire sentence. Therefore, we can obtain the factual output ${H}={\{{h_{[CLS]},h_1,...,h_M}\}}$, where ${h_{[CLS]}}$ is the global sentence feature and ${h_i}$ ($i\in\{1,...,M\}$) is the extracted word representation for $x_{i}$.
{\setlength{\abovedisplayskip}{5pt}
\setlength{\belowdisplayskip}{5pt}
\begin{equation}
    h_i=BERT(x_{i};\theta^{bert})
\end{equation}}where $\theta^{bert}$ is the BERT parameter. Particularly, if $x_i$ is split into several sub-tokens through the tokenizer, we get $h_i$ by summing the sub-tokens. 

Next, we feed the $H$ into a textual Transformer for the further encoding. As a result, we can receive the output $\{h^T_{[CLS]},h^T_1,...,h^T_M\}$

\begin{figure*}[h]
    \centering

    \includegraphics[height=0.8in,width=6.9in]{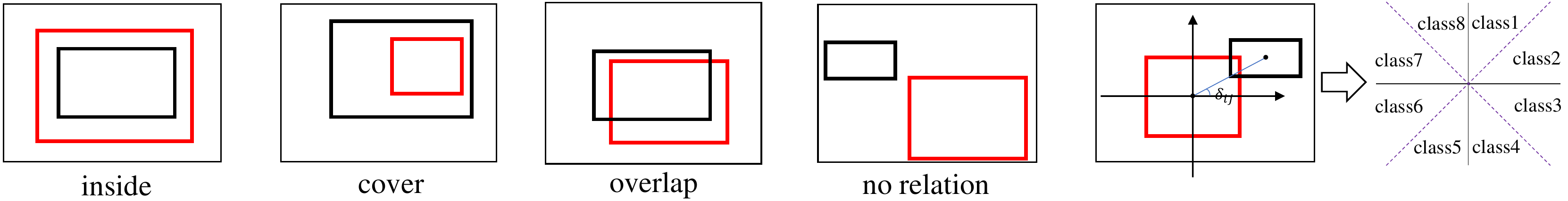}

  \caption{We have the relation between object $i$ (red region) and object $j$ (black region).}
    \label{rgcnsample}
    \vspace{-0.4cm}
  \end{figure*} 

\subsubsection{Vision Encoding}
To our intuition, image-level feature represents the global visual information (containing the scene and category, etc) of each image. But the original features directly extracted from general vision models can usually not satisfy the MNER requirement due to their specific task objectives. Following \cite{eccv2018rgcn, pami2020rgcn, vit22pami}, we enhance the global representation of the image-level by analyzing the objects and the relationship between them in the image. Also, the local representation of each object can benefit from their companions and superiors through the interaction learning.

Here, for both forms of encoding, we firstly get the initial image-level and object-level representations. Then, the encoders are applied to update the representations by capturing the relationship between different elements (i.e., including image and objects) on spatial/semantic views.

For the global representation of whole image, we choose ResNet \cite{he_resnet} as the feature extractor considering that it is one of the most excellent CNN models for many vision tasks. We take the feature map from the last convolutional layer in a pre-trained 152-layers ResNet to represent $\textbf{I}$. Then, we transform the feature map from ResNet into a one-dimensional vector by linear function. We can denote the feature vector as follows:
{\setlength{\abovedisplayskip}{5pt}
\setlength{\belowdisplayskip}{5pt}
\begin{equation}
  v_{[IMG]}=L(ResNet(\textbf{I};\theta^{res}); \theta^{I}) 
\end{equation}}where $\theta^{res}$ is the ResNet parameter and $\theta^{I}$ is the learnable parameter of linear transformation .

For local features of objects, we employ a Faster-RCNN \cite{fasterrcnn} model pre-trained on the Visual Genome \cite{vgene} to detect the objects $\{o_1, o_2, ..., o_N\}$ in the image and output the representation of each object region proposal. What's more, we adopt the concept classified by the Faster-RCNN for each object as another feature clue. Therefore, we can describe the feature representation of object $o_i$ as:{
\begin{equation}
  v_i = L(f_i;\theta^{v}) + W_ee^v_i 
\end{equation}}where $N$ indicates top-$N$ detected objects with higher scores, $f_i$ is the object region feature extracted by Faster-RCNN, $\theta^{v}$ is the learnable parameter of linear transformation, $W_e$ is the projection matrix, $e^v_i$ is the label embedding by looking up the embedding table $E^v$.

\textbf{Semantic Encoding}

As the same with \textbf{Text Encoding}, we arrange the whole image and objects into a sequence $V=\{v_{[IMG]}, v_1, v_2, ..., v_N\}$ and encourage the interaction among them to learn contextual presentations by Vision Transformer (ViT) from a semantic view. After the semantic encoding, we can get the result:{
\begin{equation}
  \{v^1_{[IMG]}, v^1_1, v^1_2, ..., v^1_N\}=ViT(V; \theta^{ViT})
\end{equation}}where $\theta^{ViT}$ is the learnable parameter of ViT.

\textbf{Spatial Encoding}

Different from the neat semantic sequence, the objects in the visual space are often scattered and irregular. We thus build the structure graph for realizing spacial modeling, resembling to the approaches of \cite{eccv2018rgcn, relation2020ocr, pami2020rgcn}.

 Firstly, we have the spatial relations between any two objects based on their region sizes and locations. For the object $o_i$, we can denote its location as $(x^c_i, y_i^c, h^c_i, w^c_i)$, where $(x^c_i, y_i^c)$ is the normalized coordinate center of $o_i$ region box, $h^c_i$ is the normalized box height and $w^c_i$ is the normalized box width. The initial spacial feature of $o_i$ is defined as $\hat{v}_i^{2}=[x^c_i, y_i^c, h^c_i, w^c_i, v_i]$. Specially, for the whole image $\textbf{I}$, we set its initial spacial feature as $\hat{v}_{[IMG]}^{2}=[x^c_{[IMG]}, y_{[IMG]}^c, h^c_{[IMG]}, w^c_{[IMG]}, v_{[IMG]}]$, where $(x^c_{[IMG]}, y_{[IMG]}^c)$, $h^c_{[IMG]}$, $w^c_{[IMG]}$ are the centroid, height, width of the image, respectively.  

Given a pair of objects $o_i$ and $o_j$, their relationship $r_{ij}$ can be depicted in the following. We compute the relative distance $d_{ij}$, Intersection over Union $u_{ij}$, and relative angle $\delta _{ij}$ (i.e., angle between vector $(x^c_j-x^c_i, y_j^c-y_i^c)$ and the positive direction of x-axis). If $o_i$ includes $o_j$ completely, we establish an edge $o_i \rightarrow o_j$ with a relation label “inside"; if conversely $o_i$ is contained by $o_j$ fully, we set the edge label as “cover". Except for the above two cases, we define the edge $o_i \rightarrow o_j$ with a label “overlap" if $u_{ij} \textgreater 0.5$. If none of these is true, we build the relationship rely on the ratio $\rho_{ij}$ (i.e., the ratio between $d_{ij}$ and the diagonal length of image) and relative angle $\delta _{ij}$. We can not have an edge for ($o_i$, $o_j$) when $\rho_{ij} \textgreater 0.5$ but assign $r_{ij}$ into one of the relationship set \{class1, class2, class3, class4, class5, class6, class7 and class8\} according to $\delta _{ij}$ when $\rho_{ij} \textless 0.5$. (Please see Figure \ref{rgcnsample} for more details.) By default, we think the $\textbf{I}$ includes all the objects.

The spatial structure graph $G$=$(\mathbb{V}, \mathbb{E})$ is then derived through the object-to-object relations, where $\mathbb{V}=\{\textbf{I}, o_1, ..., o_N\}$ is the node set, $\mathbb{E}$ is the established edge set with relations. Imitating the operation in text \cite{supernode}, we regard the $\textbf{I}$ in the graph as a super node to collect the global visual information. Next, we update the node representations by using a modified R-GCN with the initial spacial features.

For the vertex $i \in \mathbb{V}$ in $k$ layer of GCN, the accumulating information $v_i^{2,k'}$ from its neighbors can be formalized as follow with considering the relationship label and direction of all connection edges: {\setlength{\abovedisplayskip}{3pt}
\setlength{\belowdisplayskip}{3pt}
\begin{equation}
  v_i^{2,k'}=\phi (\sum_{j \in \Omega (i)}{W_{r_{i \leftrightarrow j}}v_j^{2,k}+b})
\end{equation}}where $\phi $ is a nonlinear function, $\Omega (i)$ is the neighbor set of $i$, $r_{i \leftrightarrow j}$ denotes the directional relationship between $i$ and $j$, $W_{r_{i \leftrightarrow j}}$ indicates the transformation weight with regard to the edge direction and label, $v_j^{2,k}$ is the representation of $j$ in $k$ layer, $b$ is the bias term. 

We update the representation of $i$ in $k+1$ layer through a gate mechanism:{\setlength{\abovedisplayskip}{3pt}
\setlength{\belowdisplayskip}{1.5pt}
\begin{equation}
  \lambda_{i,g}^{2,k}=sigmoid(W_{gcn}[v_i^{2,k'}, v_i^{2,k}])
  \end{equation}
  \begin{equation}
  v_{i}^{2,k+1}=v_i^{2,k}+\lambda_{i,g}^{2,k}v_i^{2,k'}
\end{equation}}where sigmoid($\cdot$) is the activation function, $W_{gcn}$ is the trainable matrix.

After the spacial encoding and learning among graph nodes, we can finally have the local and global visual feature vectors $\{v^2_{[IMG]}, v^2_1, v^2_2, ..., v^2_N\}$.

\subsection{Relevance Measuring}

Unlike the current works that compute the match degree between image and text relying on the extra image-text classification task and VLM, we dynamically measure the relevance only depend on the MNER objective. We can define the relevance score $M^r$ ($r\in\{1,2\}$) between global text feature $h^T_{[CLS]}$ and vision feature $v^r_{[IMG]}$ as:{\setlength{\abovedisplayskip}{4pt}
\setlength{\belowdisplayskip}{4pt}
\begin{equation}
  C^r=tanh(h^T_{[CLS]}W^r_{TI}v^r_{[IMG]})
\end{equation}
\begin{equation}
  M^r=tanh(W^r_Th^T_{[CLS]}+W^r_Iv^r_{[IMG]}C^r)
\end{equation}}where $W^r_{TI}$, $W^r_T$ and $W^r_I$ are the learnable weight matrices, tanh($\cdot$) is the activation function. Based on the measure result, we get the local-global vision feature $\textbf{V}^r$ as follows:{
\begin{equation}
  v^r_i = L([M^rv^r_{[IMG]}, v^r_i]; \theta^M), i\in\{1,2,..N\}
  \end{equation}
\begin{equation}
\textbf{V}^r=[v^r_1,...,v^r_N] , r\in\{1,2\}
\end{equation}}where $\theta^M$ is the parameter of linear function.
\subsection{Inter-modality Learning}
During this stage, different modalities iteratively encourage each other to acquire the most powerful multimodal feature. As the Figure \ref{model_overview} shows, this interaction module contains three parts which all take the cross-modal Transformer as the core encoder. To reduce the heterogeneity between text and vision, we transform the text representation before the cross-modal learning:{
\begin{equation}
  \textbf{H}=FFNN(\{h^T_1,...,h^T_M\}; \theta^{FFNN})
\end{equation}}where $\theta^{FFNN}$ is a parameter for FFNN training.

Subsequently, the detailed illustrations of theses three parts are given (top-down). At the first part, we refine representations of textual words by introducing the fused spatial and semantic visual information. The comprehensive vision feature can be resulted by a gate control:{
\begin{equation}
  \alpha = W^V\sigma(W^{V1}\textbf{V}^1+W^{V2}\textbf{V}^2)
\end{equation}
\begin{equation}
  \textbf{V} = \alpha \odot \textbf{V}^1+(\boldsymbol{1}-\alpha) \odot \textbf{V}^2
\end{equation}}where $W^V$, $W^{V1}$ and $W^{V2}$ are the trainable matrices, $\sigma(\cdot)$ is the activation function, $\odot$ represents element-wise multiplication, $\boldsymbol{1}$ stands for an all-1 vector. For inputs $(Q,K,V)$ of the Transformer, we assign $\textbf{H}$ to $Q$, $\textbf{V}$ to $(K,V)$ for the update of $\textbf{H}$. That is:{
\begin{equation}
  \textbf{H} = Transformer(\textbf{H}, \textbf{V}; \theta^{FT})
\end{equation}}where $\theta^{FT}$ is the learnable parameter for final multimodal text encoding.

Similar to the first part, we can update $\textbf{V}^1$ through feeding $\textbf{V}^1$ to $Q$, $\textbf{H}$ to $(K,V)$ in the second part.
Also in the third part, we update $\textbf{V}^2$ by taking $\textbf{V}^2=Q$, $\textbf{H}=K=V$.

After $\textbf{\textit{L}}$ synchronous iterations of all parts, we use the final $\textbf{H}$ for MNER decoding.

\subsection{MNER Decoding}

Following \cite{moon2018multimodal}, we produce the probability of a predicted label sequence $y$ by feeding $\textbf{H}$ to a CRF layer:
{\setlength{\abovedisplayskip}{4pt}
\setlength{\belowdisplayskip}{4pt}
  {
\begin{eqnarray}
p(y|\textbf{H};\theta^{CRF})=\frac{\prod^{M-1}_{i=1}{\varphi_i(y_i,y_{i+1};\textbf{H})}}{\sum_{y'\in {\mathbb{Y}}}\prod^{M-1}_{i=1}{\varphi_i({y'}_i,{y'}_{i+1};\textbf{H})}}
\end{eqnarray}}}where $\varphi_i(y_i,y_{i+1};\textbf{H})$ is a potential function, ${\mathbb{Y}}$ is a set of all possible label sequences, $\theta^{CRF}$ is a set of parameters which define the potential function and the transition score from the label $y_i$ to the label $y_{i+1}$. 

We train the model by maximizing conditional likelihood estimation for the training set $\{(\textbf{X},\textbf{Y})_t\}$:
{\setlength{\abovedisplayskip}{3pt}
\setlength{\belowdisplayskip}{3pt}
\begin{eqnarray}
Loss=\sum_{t}log\ p(\textbf{Y}|\textbf{H};\theta^{CRF})
\end{eqnarray}}

In the decoding phase, we output the label sequence prediction $y^*$ for given \textbf{X} based on maximizing the following score:
{\setlength{\abovedisplayskip}{3pt}
\setlength{\belowdisplayskip}{3pt}
\begin{equation}
y^* = \mathop{\arg\max}_{y\in \mathbb{Y}} p(y|\textbf{H};\theta^{CRF})
\end{equation}}

\section{Experiments}

We evaluate our model on two publicly MNER datasets referring to \cite{yu2020improving,ita2022,Chen2022HybridTW,catmner2022} and compare it with a number of approaches.

\begin{table}[t]
 
 
  \centering

  {\fontsize{8.5pt}{9.5pt}\selectfont
  \begin{tabular}{l|lll|lll}
  \toprule
  \multirow{2}{*}{Ent Type} &
    \multicolumn{3}{c|}{TWITTER-2015} &
    \multicolumn{3}{c}{TWITTER-2017} \\ \cline{2-7} 
   &
    \multicolumn{1}{c}{Train} &
    \multicolumn{1}{c}{Val} &
    \multicolumn{1}{c|}{Test} &
    \multicolumn{1}{c}{Train} &
    \multicolumn{1}{c}{Val} &
    \multicolumn{1}{c}{Test} \\ \hline
  PER        & 2217 & 552  & 1816 & 2943 & 626  & 621  \\
  LOC      & 2091 & 522  & 1697 & 731  & 173  & 178  \\
  ORG  & 928  & 247  & 839  & 1674 & 375  & 395  \\
  MISC & 940  & 225  & 726  & 701  & 150  & 157  \\ \hline
  Total       & 6176 & 1546 & 5078 & 6049 & 1324 & 1351 \\ \hline
  Tweets & 4000 & 1000 & 3257 & 3373 & 723  & 723  \\ 
  \bottomrule
  \end{tabular}}
    \caption{The basic statistics for both datasets.}
    \label{tabdata}
    \vspace{-0.4cm}
  \end{table}

\subsection{Datasets}
The experiments are carried out on the datasets TWITTER-2015 and TWITTER-2017, which are constructed based on Twitter by \cite{lu2018visual} and \cite{zhang2018adaptive} separately. TWITTER-2015 contains 12800 entities and the number of tweets is 8257. TWITTER-2017 contains 8724 entities and the total number of tweets is 4819. For the fairness of comparison, we take the same split with previous works (4000 for training, 3257 for test, and 1000 for validation) in TWITTER-2015, (3373 for training, 723 for test, 723 for validation) in TWITTER-2017, respectively. Table \ref{tabdata} summarizes the two sizes.

\begin{table*}[]

{
\fontsize{6.88pt}{7.0pt}\selectfont
\centering

\begin{tabular}{c|c|cllllll|lllllll}
\toprule 
 &  & \multicolumn{7}{c|}{TWITTER-2015} & \multicolumn{7}{c}{TWITTER-2017} \\ \hline
 \multirow{2}{*}{Modality}& \multirow{2}{*}{Methods} & \multicolumn{4}{c|}{Single Type (F1)} & \multicolumn{3}{c|}{Overall} & \multicolumn{4}{c|}{Single Type (F1)} & \multicolumn{3}{c}{Overall} \\
 &  & PER & \multicolumn{1}{c}{LOC} & \multicolumn{1}{c}{ORG} & \multicolumn{1}{c|}{MISC} & \multicolumn{1}{c}{P} & \multicolumn{1}{c}{R} & \multicolumn{1}{c|}{F1} & \multicolumn{1}{c}{PER} & \multicolumn{1}{c}{LOC} & \multicolumn{1}{c}{ORG} & \multicolumn{1}{c|}{MISC} & \multicolumn{1}{c}{P} & \multicolumn{1}{c}{R} & \multicolumn{1}{c}{F1} \\ \hline
 & BiLSTM-CRF & {\color[HTML]{333333} 76.77} & \multicolumn{1}{c}{{\color[HTML]{333333} 72.56}} & \multicolumn{1}{c}{{\color[HTML]{333333} 41.33}} & \multicolumn{1}{c|}{{\color[HTML]{333333} 26.80}} & \multicolumn{1}{c}{{\color[HTML]{333333} 68.14}} & \multicolumn{1}{c}{{\color[HTML]{333333} 61.09}} & \multicolumn{1}{c|}{{\color[HTML]{333333} 64.42}} & \multicolumn{1}{c}{{\color[HTML]{333333} 85.12}} & \multicolumn{1}{c}{{\color[HTML]{333333} 72.68}} & \multicolumn{1}{c}{{\color[HTML]{333333} 72.50}} & \multicolumn{1}{c|}{{\color[HTML]{333333} 52.56}} & \multicolumn{1}{c}{{\color[HTML]{333333} 79.42}} & \multicolumn{1}{c}{{\color[HTML]{333333} 73.43}} & \multicolumn{1}{c}{{\color[HTML]{333333} 76.31}} \\
 & CNN-BiLSTM-CRF & {\color[HTML]{333333} 80.86} & \multicolumn{1}{c}{{\color[HTML]{333333} 75.39}} & \multicolumn{1}{c}{{\color[HTML]{333333} 47.77}} & \multicolumn{1}{c|}{{\color[HTML]{333333} 32.61}} & \multicolumn{1}{c}{{\color[HTML]{333333} 66.24}} & \multicolumn{1}{c}{{\color[HTML]{333333} 68.09}} & \multicolumn{1}{c|}{{\color[HTML]{333333} 67.15}} & \multicolumn{1}{c}{{\color[HTML]{333333} 87.99}} & \multicolumn{1}{c}{{\color[HTML]{333333} 77.44}} & \multicolumn{1}{c}{{\color[HTML]{333333} 74.02}} & \multicolumn{1}{c|}{{\color[HTML]{333333} 60.82}} & \multicolumn{1}{c}{{\color[HTML]{333333} 80.00}} & \multicolumn{1}{c}{{\color[HTML]{333333} 78.76}} & \multicolumn{1}{c}{{\color[HTML]{333333} 79.37}} \\
 & HBiLSTM-CRF & {\color[HTML]{333333} 82.34} & \multicolumn{1}{c}{{\color[HTML]{333333} 76.83}} & \multicolumn{1}{c}{{\color[HTML]{333333} 51.59}} & \multicolumn{1}{c|}{{\color[HTML]{333333} 32.52}} & \multicolumn{1}{c}{{\color[HTML]{333333} 70.32}} & \multicolumn{1}{c}{{\color[HTML]{333333} 68.05}} & \multicolumn{1}{c|}{{\color[HTML]{333333} 69.17}} & \multicolumn{1}{c}{{\color[HTML]{333333} 87.91}} & \multicolumn{1}{c}{{\color[HTML]{333333} 78.57}} & \multicolumn{1}{c}{{\color[HTML]{333333} 76.67}} & \multicolumn{1}{c|}{{\color[HTML]{333333} 59.32}} & \multicolumn{1}{c}{{\color[HTML]{333333} 82.69}} & \multicolumn{1}{c}{{\color[HTML]{333333} 78.16}} & \multicolumn{1}{c}{{\color[HTML]{333333} 80.37}} \\ \cline{2-16} 
 & BERT & {\color[HTML]{333333} 84.72} & \multicolumn{1}{c}{{\color[HTML]{333333} 79.91}} & \multicolumn{1}{c}{{\color[HTML]{333333} 58.26}} & \multicolumn{1}{c|}{{\color[HTML]{333333} 38.81}} & \multicolumn{1}{c}{{\color[HTML]{333333} 68.30}} & \multicolumn{1}{c}{{\color[HTML]{333333} 74.61}} & \multicolumn{1}{c|}{{\color[HTML]{333333} 71.32}} & \multicolumn{1}{c}{{\color[HTML]{333333} 90.88}} & \multicolumn{1}{c}{{\color[HTML]{333333} 84.00}} & \multicolumn{1}{c}{{\color[HTML]{333333} 79.25}} & \multicolumn{1}{c|}{{\color[HTML]{333333} 61.63}} & \multicolumn{1}{c}{{\color[HTML]{333333} 82.19}} & \multicolumn{1}{c}{{\color[HTML]{333333} 83.72}} & \multicolumn{1}{c}{{\color[HTML]{333333} 82.95}} \\
\multirow{-5}{*}{Text} & BERT-CRF & \multicolumn{1}{l}{84.74} & 80.51 & 60.27 & \multicolumn{1}{l|}{37.29} & 69.22 & 74.59 & 71.81 & 90.25 & 83.05 & 81.13 & \multicolumn{1}{l|}{62.21} & 83.32 & 83.57 & 83.44 \\ \hline

 & GVATT-BERT-CRF & \multicolumn{1}{l}{84.43} & 80.87 & 59.02 & \multicolumn{1}{l|}{38.14} & 69.15 & 74.46 & 71.70 & 90.94 & 83.52 & 81.91 & \multicolumn{1}{l|}{62.75} & 83.64 & 84.38 & 84.01 \\

 & AdaCAN-BERT-CRF & \multicolumn{1}{l}{85.28} & 80.64 & 59.39 & \multicolumn{1}{l|}{38.88} & 69.87 & 74.59 & 72.15 & 90.20 & 82.97 & 82.67 & \multicolumn{1}{l|}{64.83} & 85.13 & 83.20 & 84.10 \\
 
 
 & UMT & \multicolumn{1}{l}{85.24} & 81.58 & 63.03 & \multicolumn{1}{l|}{39.45} & 71.67 & 75.23 & 73.41 & {91.56} & 84.73 & 82.24 & \multicolumn{1}{l|}{{70.10}} & 85.28 & 85.34 & 85.31 \\    
& UMGF & \multicolumn{1}{l}{84.26} & \textbf{83.17} & 62.45 & \multicolumn{1}{l|}{42.42} & {74.49} & 75.21& 74.85 & {91.92} & 85.22 & 83.13 & \multicolumn{1}{l|}{{69.83}} & 86.54 & 84.50 & 85.51 \\ 

& GEI & \multicolumn{1}{l}{-} & \textbf{-} & - & \multicolumn{1}{l|}{-} & {73.39  } & 75.51& 74.43 & {-} & - & - & \multicolumn{1}{l|}{-} & 87.50   & 86.01 & 86.75 \\

& MAF & \multicolumn{1}{l}{84.67} & 81.18 & {63.35} & \multicolumn{1}{l|}{{41.82}} & 71.86 & 75.10 & 73.42 & 91.51 & 85.80 & 85.10 & \multicolumn{1}{l|}{68.79} & {86.13} & {86.38} & {86.25} \\

& RDS & \multicolumn{1}{l}{-} & - & {-} & \multicolumn{1}{l|}{-} & 71.96   & 75.00 & 73.44 & - & - & - & \multicolumn{1}{l|}{-} & {86.25 
} & {86.38 } & {86.32} \\

& ITA & \multicolumn{1}{l}{85.6} & 82.6 & \underline{64.4} & \multicolumn{1}{l|}{\textbf{44.8}} & - & - & \underline{75.60} & 91.4 & 84.8 & 84.0 & \multicolumn{1}{l|}{68.6} & - & - & 85.72 \\
& MRC-MNER & \multicolumn{1}{l}{\underline{85.71}} & 81.97 & 61.12 & \multicolumn{1}{l|}{40.20} & \textbf{78.10} & 71.45 & 74.63 & \underline{92.64} & \textbf{86.47} & 83.16 & \multicolumn{1}{l|}{\textbf{72.66}} & \textbf{88.78} & 85.00 &86.85 \\

& MNER-QG & \multicolumn{1}{l}{85.31 } & 81.65  & 63.41  & \multicolumn{1}{l|}{41.32} & \underline{77.43 } & 72.15  & 74.70 & \textbf{92.92 } & {86.19 } & 84.52  & \multicolumn{1}{l|}{\underline{71.67}} & \underline{88.26 } & 85.65  &\underline{86.94} \\

& MGCMT & \multicolumn{1}{l}{\textbf{85.84}} & 82.03 & 63.08 & \multicolumn{1}{l|}{40.81} & 73.57 & 75.59 & 74.57 & 90.82 & 86.21 & \underline{86.26} & \multicolumn{1}{l|}{66.88} & 86.03 & 86.16 & 86.09 \\
& HVPNeT & \multicolumn{1}{l}{-} & - & - & \multicolumn{1}{l|}{-} & 73.87 & \textbf{76.82} & 75.32 & - & - & - & \multicolumn{1}{l|}{-} & 85.84 & \textbf{87.93} & 86.87 \\
\cline{2-16}


\multirow{-12}{*}{\begin{tabular}[c]{@{}l@{}}Text+\\ Image \end{tabular}} & \textbf{HamLearning} & {{85.28}} & \multicolumn{1}{c}{{\underline{82.84}}} & \multicolumn{1}{c}{\textbf{64.46}} & \multicolumn{1}{c|}{\underline{42.52}} & \multicolumn{1}{c}{{{77.25}}} & \multicolumn{1}{c}{{\underline{75.75}}} & \multicolumn{1}{c|}{{\textbf{76.49}}} & {{91.43}} & {\underline{86.26}} & {\textbf{86.66}} & \multicolumn{1}{l|}{{{69.17}}} & {86.99} & \underline{87.28} & \textbf{87.13} \\ 
\bottomrule
\end{tabular}}
\caption{Performance comparison of different competitive uni-modal and multi-modal approaches.}
\label{mainresult}
\vspace*{-0.3cm}
\end{table*}

\subsection{Implementation Details}
For both datasets, we have the same hyperparameters. In the experiments, the batch size of training is 32 while it is 16 during validation and test, and the maximum length of the input text sequence is 128 which can cover all words. The initial representations ${H}$ are encoded with the uncased \textit{BERT$_{base}$} model pre-trained by \cite{devlin2019bert} with the dimension of 768. The feature dimension of whole image and objects after linear transformation is 768, the number of ViT layers is 4 and the number of R-GCN layers is 2. The head size in multi-head attention is 12, the feature dimensions in all the Transformers are 768. The dropout rate, the learning rate, the number of detected objects and epochs are respectively set to 0.1, 3e-5, 15, 60. The number of FFNN layers is 2, and we test the number $\textbf{\textit{L}}$ to find the best. We perform our experiments on the Tesla-V100 GPU. 

\subsection{Baselines}We compare our model with some typical excellent approaches for NER, including unimodal approaches (only text as inputs) and multimodal approaches (text-image pairs as inputs). For unimodal approaches, we consider: \textbf{BiLSTM-CRF} \cite{huang2015bidirectional}. \textbf{CNN-BiLSTM-CRF} \cite{ma-hovy-2016-end}, extends the work of BiLSTM-CRF and incorporates the character-level word representation learned by CNN into input. \textbf{HBiLSTM-CRF} \cite{lample2016neural}, similar to CNN-BiLSTM-CRF, but get the character-level word representation from LSTM. \textbf{BERT} \cite{devlin2019bert} and its variant \textbf{BERT+CRF}. For multimodal approaches, we consider: \textbf{AdaCAN-BERT-CRF} \cite{zhang2018adaptive} and \textbf{GAVTT-BERT-CRF} \cite{lu2018visual}, which combine the whole image feature through visual attention, and we replace their original sentence encoders BiLSTM with BERT. \textbf{UMT} \cite{yu2020improving} creates the cross-modal Transformer to encode the image region and text for MNER. \textbf{MAF} \cite{xu2022maf} and \textbf{RDS} \cite{extradata}, design the extra contrastive tasks to make text and image more consistent and assign visual feature to assist words. \textbf{UMGF} \cite{Zhang2021UMGF} and \textbf{GEI} \cite{umgf2}, which enhance cross-modal interaction between visual objects and textual words with GNN. \textbf{MRC-MNER} \cite{MRC2022MM}, \textbf{MNER-QG} \cite{querymultivision23ai} and \textbf{ITA} \cite{ita2022} which leverage the prior knowledge, caption and OCR of the image to guide vision-aware word information. \textbf{MGCMT} \cite{Liu2022MultiGranularityCR} and \textbf{HVPNeT} \cite{chen2022HVPM}, the initial attempts for multi-level semantic alignment at different vision levels.

\begin{figure}[h]
  \centering
  \includegraphics[height=1.3in,width=3.35in]{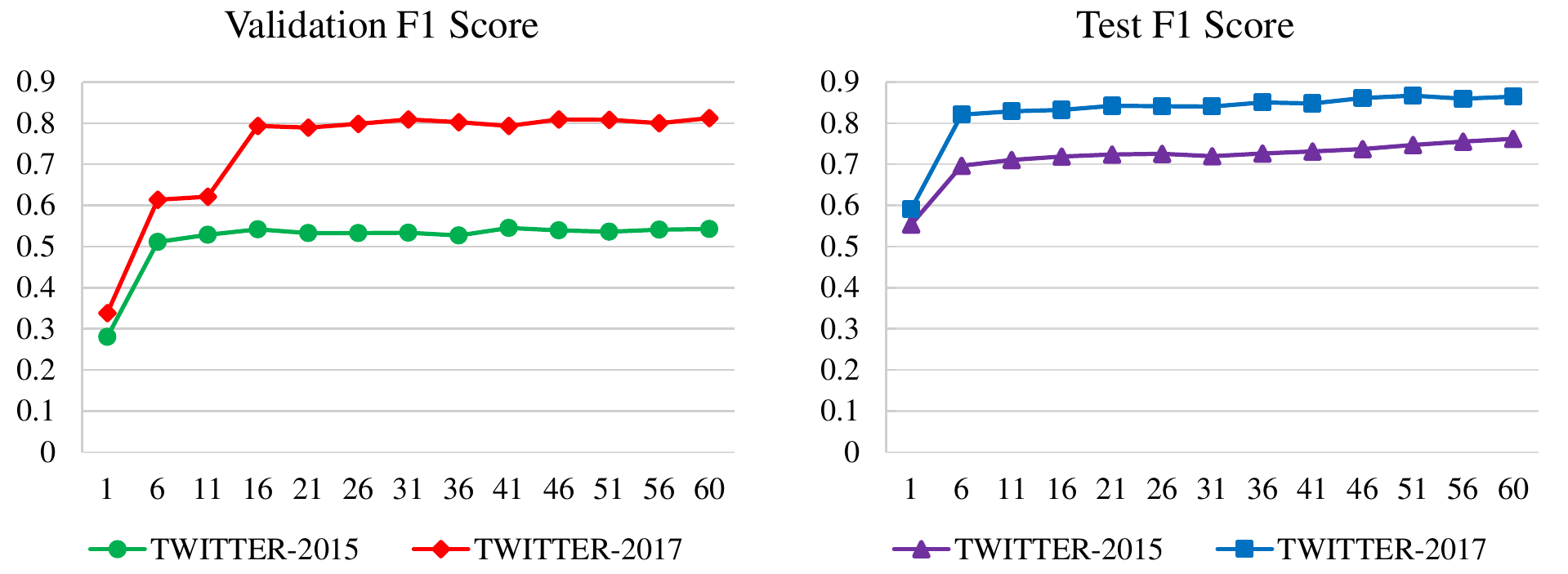}
  \caption{The changes of important indicators (i.e., Loss and F1) during the training process of our model.}  
  \label{model_excet}
\end{figure}

\subsection{Main Results}

  Following the literature \cite{extradata,umgf2, chen2022HVPM}, we compute experimental results of F1 score (\textbf{F1}) for every single type and overall precision (\textbf{P}), recall (\textbf{R}), and F1 score (\textbf{F1}). For a fair comparison, we refer to the results of all baselines introduced in their papers. Figure \ref{model_excet} depicts the training process of our model, and Table \ref{mainresult} shows all comparison results. The results suggest that both overall F1s of our method on the two benchmark datasets outperform the published state-of-the-art (SOTA) performance. We also have several findings below:

  (1) It is clear that BERT-based methods perform better compared with BiLSTM-based encoders ((BiLSTM-CRF, CNN-BiLSTM-CRF) vs (BERT, BERT-CRF)), which indicates that the pre-trained model is quite effective due to its large external knowledge support. 
  
  (2) Through all multimodal and unimodal approaches, we can see that unimodal approaches generally produce poor performance compared to the multimodal ones, which suggests that visual information of either the global image or the local objects is valuable for MNER. Comparing UMT and UMGF/GEI, we find that UMGF/GEI can get the better result than UMT, which may be that more entities can be guided by fine-grained visual objects. With the performances of UMT and MAF/RDS, we can notice the importance of evaluating the text-image matching degree and eliminating the vision noise. 
  
  (3) As the results of ITA, MRC-MNER and MNER-QG show, transforming and guiding the image content into text knowledge to assist NER can also be effective. However, in fact, this method depends on the effect of external energies such as OCR model and hand-crafted templates. 
  
  (4) From HamLearning, HVPNeT, MGCMT and others, we can observe that the multi-level semantic interaction is usually more advantageous than single grained information. Considering HamLearning and HVPNeT/MGCMT, we probably think that the more comprehensive and in-depth interaction could lead to better results since HVPNeT and MGCMT focus on the multi-level vision while HamLearning performs on both modalities.

  (5) The performance of MAF/RDS and HamLearning presents us the difference of two text-image matching methods, we believe that the dynamic measuring is more reasonable and reliable because of the direct connection of end-to-end to the specific MNER task instead of the parallel classification task. 

  \subsection{Ablation Study}

  To investigate the contribution of main modules in our model, we have an ablation study. Table \ref{ablation} reports comparison results between the full model and its ablation methods. (For convenience, we here use R-GCN to represent the spatial visual encoding, ViT to represent the semantic visual encoding, and RM to represent the relevance measuring.) We find that: 1) All the modules have contributions to our final optimal model, and any removal of the three modules would result in the inferior performance. 2) When we remove the R-GCN, the hidden information behind the image including the scene knowledge and the relationship between objects will be difficult to be captured for multimodal reasoning, and we only use the semantic feature extracted by ViT as the visual supplement. As the results show, the R-GCN plays an more important role compared to the ViT. 3) From the table, we observe that the elimination of the relevance measuring leads to the significant performance degradation. Especially, the recall on both datasets drop a lot. The reason may be that, with the removal of relevance measuring, a large amount of redundant visual feature has produced misleading noise, which damages the ability of the model to detect the correct entities. Furthermore, as described in our method, the absence of RM can also affect the role of R-GCN and ViT. 

  \begin{table}[]

\fontsize{7.8pt}{9pt}\selectfont
\centering

{\begin{tabular}{l|lll|lll}
\toprule
\multicolumn{1}{c|}{\multirow{2}{*}{Settings}} & \multicolumn{3}{c|}{TWITTER-2015}                                      & \multicolumn{3}{c}{TWITTER-2017}                                      \\ \cline{2-7} 
\multicolumn{1}{c|}{}                         & \multicolumn{1}{c}{P} & \multicolumn{1}{c}{R} & \multicolumn{1}{c|}{F} & \multicolumn{1}{c}{P} & \multicolumn{1}{c}{R} & \multicolumn{1}{c}{F} \\ \hline
\multicolumn{1}{c|}{Default}                  & 
      \textbf{77.25} &     75.75 &   \textbf{76.49} & \multicolumn{1}{c}{{ \textbf{86.99}}} & { \textbf{87.28}} & { \textbf{87.13}}                  \\ \hline
w/o  RGCN          & 74.73             & 75.42          & 75.07      & 86.17         & 85.94        & 86.05       \\
w/o  ViT            & {74.45}     & \textbf{75.92}         & 75.18       & 85.75         & {86.53}    & 86.14                 \\
w/o  RM            & 75.35              & 73.89        & 74.61       & 86.02         & 85.19         & 85.60     \\ \bottomrule
\end{tabular}
}
\caption{The ablation study for different module of HamLearning.}
\label{ablation}
\end{table}

\begin{table}[]

\fontsize{7.8pt}{9pt}\selectfont

\centering

{
\begin{tabular}{c|ccc|ccc}
\toprule
\multirow{2}{*}{$\textbf{\textit{L}}$}           & \multicolumn{3}{c|}{TWITTER-2015}                                      & \multicolumn{3}{c}{TWITTER-2017}                                      \\ \cline{2-7} 
                                    & \multicolumn{1}{c}{P} & \multicolumn{1}{c}{R} & \multicolumn{1}{c|}{F} & \multicolumn{1}{c}{P} & \multicolumn{1}{c}{R} & \multicolumn{1}{c}{F} \\ \hline
\multicolumn{1}{c|}
{\iffalse obj-num=4 \fi     1} &
  74.97 &
  75.49 &
  75.23 &
  85.30 &
  \multicolumn{1}{c}{86.16} &
  \multicolumn{1}{c}{85.73} \\
  2 &
  {{76.61}} &
  { {75.36}} &
  { {75.98}} &
  \multicolumn{1}{c}{86.97} &
  86.38 &
  86.67 \\
  3 &
  \textbf{77.25} &
  75.75 &
  \textbf{76.49} &
  \multicolumn{1}{c}{{ {86.99}}} &
  { \textbf{87.28}} &
  { \textbf{87.13}} \\
  4 &
  {76.33} &
  75.21 &
  75.77 &
  \multicolumn{1}{c}{\textbf{87.01}} &
  85.79 &
  86.40 \\ 
  5 &
  75.25 &
  \textbf{76.03} &
  75.64 &
  \multicolumn{1}{c}{{ {85.43}}} &
  { {86.57}} &
  { {86.00}} \\
  \bottomrule
\end{tabular}
}

\caption{The performance of HamLearning by different $\textbf{\textit{L}}$ numbers.}
\label{tabl_num}
\end{table}

\subsection{Further Analysis}

\subsubsection{Parameter Sensitivity}  
In this section, we evaluate our model on different number $\textbf{\textit{L}}$ to find the optimal parameter. The Table \ref{tabl_num} shows the experiment results. As the results show, with the increase of $\textbf{\textit{L}}$, the performance of the model becomes better. When the number is 3, we can obtain the best model. However, when the number is greater than 3, the performance begins to decline. This may be because, with the deepening of cross-modal interaction learning, the differences between modal information are decreasing, leading to the lack of valuable features.

\subsubsection{Generalization Analysis}
Considering the different data characteristics of the two datasets, we conducted cross validation on them to test the generalization ability of our model and the comparison methods. As shown in Table \ref{genablity}, TWITTER-17{$\rightarrow$}TWITTER-15 indicates that the model trained on TWITTER-2017 dataset is used to test the TWITTER-2015 dataset, and vice versa. From the results, we can discover that, our model significantly outperforms its comparisons by a large margin. This phenomenon may potentially confirm the transfer and adaptability ability of multimodal hierarchical semantics in model generalization.

\begin{table}[]

  \fontsize{6.6pt}{8pt}\selectfont
  
  \centering

  {
  \begin{tabular}{c|ccc|ccc}
  \toprule
  \multirow{2}{*}{{Methods}}           & \multicolumn{3}{c|}{TWITTER-17{$\rightarrow$}TWITTER-15}                                      & \multicolumn{3}{c}{TWITTER-15{$\rightarrow$}TWITTER-17}                                   \\ \cline{2-7} 
                                    & \multicolumn{1}{c}{P} & \multicolumn{1}{c}{R} & \multicolumn{1}{c|}{F} & \multicolumn{1}{c}{P} & \multicolumn{1}{c}{R} & \multicolumn{1}{c}{F} \\ \hline
  \multicolumn{1}{c|}
  {\iffalse obj-num=4 \fi     UMT{\ddag} } &
  64.67 &
  63.59 &
  64.13 &
  67.80 &
  \multicolumn{1}{c}{55.23} &
  \multicolumn{1}{c}{60.87} \\
  UMGF{\ddag} &
  {{67.00}} &
  { {62.81}} &
  { {66.21}} &
  \multicolumn{1}{c}{69.88} &
  56.92 &
  62.74 \\
  HamLearning &
  \textbf{69.17} &
  \textbf{ 66.84} &
  \textbf{67.98} &
  \multicolumn{1}{c}{{ \textbf{71.03}}} &
  { \textbf{59.40}} &
  { \textbf{64.70}} \\
  \bottomrule
  \end{tabular}}
    \caption{The performance comparison of generalization ability between \textbf{HamLearning} and other methods. Results with {\ddag}  are from \cite{Zhang2021UMGF}.}

  \label{genablity}
  \end{table}

  \begin{table}[]
  
\fontsize{7.5pt}{9pt}\selectfont
  
  \centering

  {
  \begin{tabular}{c|ccc|ccc}
  \toprule
  \multirow{2}{*}{{Detectors}}           & \multicolumn{3}{c|}{TWITTER-2015}                                      & \multicolumn{3}{c}{TWITTER-2017}                                   \\ \cline{2-7} 
                                    & \multicolumn{1}{c}{P} & \multicolumn{1}{c}{R} & \multicolumn{1}{c|}{F} & \multicolumn{1}{c}{P} & \multicolumn{1}{c}{R} & \multicolumn{1}{c}{F} \\ \hline
  \multicolumn{1}{c|}
{\iffalse obj-num=4 \fi     Faster RCNN } &
  {77.25} &
  75.75 &
  {76.49} &
  \multicolumn{1}{c}{{ {86.99}}} &
  {  {87.28}} &
  {  {87.13}}  \\
  Mask RCNN &
  {{76.93}} &
  { {75.74}} &
  { {76.33}} &
  \multicolumn{1}{c}{87.03} &
  86.45 &
  86.73 \\
  \bottomrule
  \end{tabular}}
  
  \caption{Performance effect from the different object detectors.}
  \label{difobj}
  \end{table}

\subsubsection{Different Object Detectors}
To explore the impact of different object detectors on model performance, we also apply Mask RCNN \cite{mrcnn_he} pre-trained on the MS COCO \cite{mscoco} dataset to detect a set of objects from the image. In fact, the MS COCO has fewer object categories than the Visual Genome. So, compared to Faster RCNN, there are fewer objects obtained from Mask RCNN leading to the slightly inferior performance (see Table \ref{difobj}). However, the performance deviation is small, indicating that the main objects and their associated scene in the image are the dominant information for extracting entities.

\begin{figure}[h]
  \centering

  \includegraphics[height=1.6in,width=3.3in]{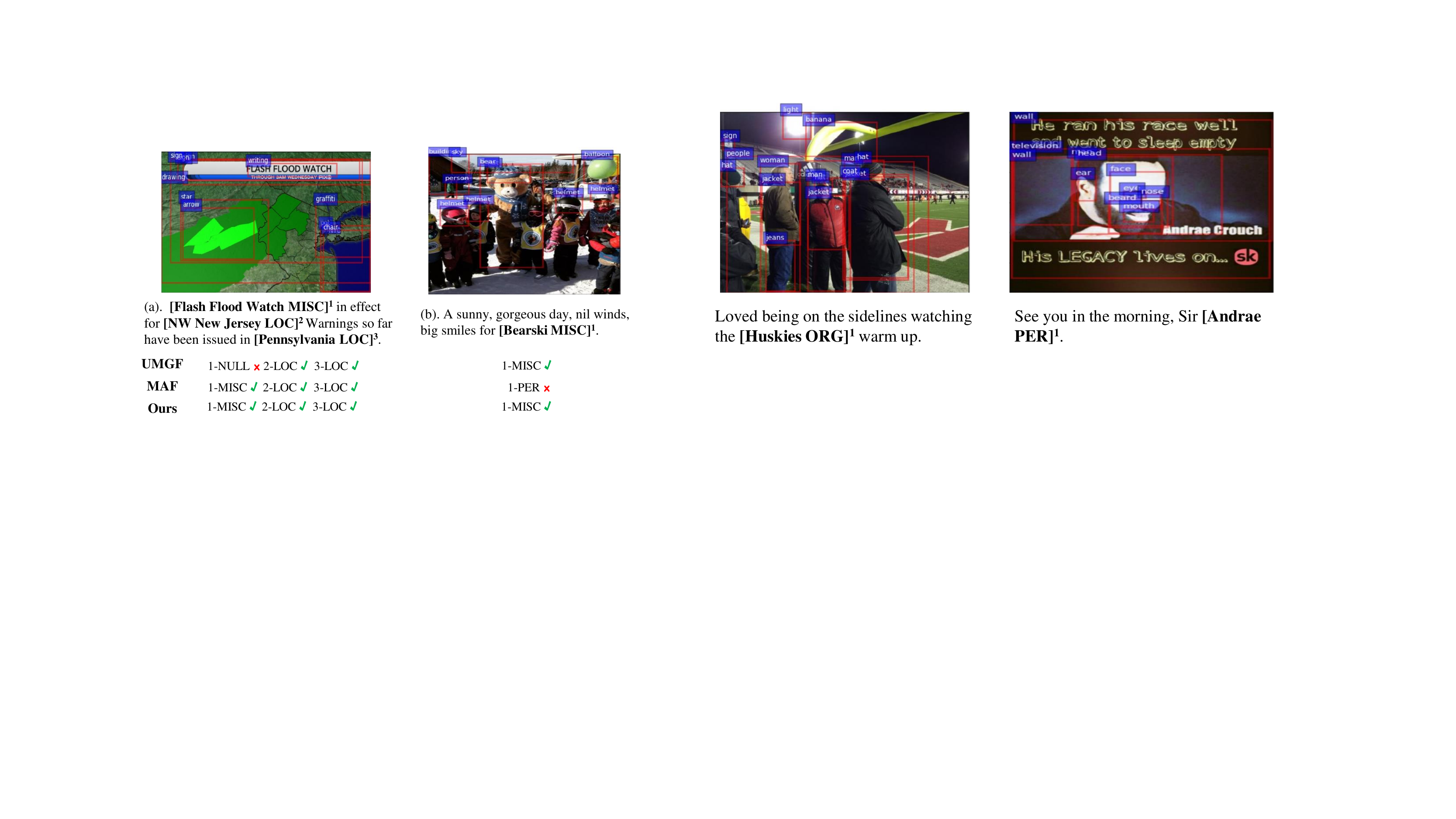}\\
  \caption{The case comparisons of our model and others.}  
\label{casestudy}
\end{figure}

\subsection{Case Study}
To better appreciate the advance of our model, we choose 2 representative test cases from the TWITTER-2015, and compare their predicted results of our model, UMGF and MAF in Figure \ref{casestudy}. We will discuss each case in the following.

As we can see, the image of case-a can not actually provide the text with effective visual guidance information (whether the global image or the detected objects) for NER. Our model and MAF prevent the impact of noisy visual information successfully. Unfortunately, UMGF has no capacity to do this without any useful objects. 

Case-b reveals the effort of the fine-grained semantic interaction between text and local objects. We can find that UMGF and our approach can accurately predict “Bearski" with the guidance of detected objects “bear" in the image while MAF obtains the wrong prediction because of the noisy guidance of the global image with many people.

\section{Conclusion}
In this paper, we introduce a novel hierarchical neural network HamLearning to achieve the multimodal learning for MNER. Our model contains three main modules: intra-modality learning, image-text relevance measuring and iterative cross-modal learning. The intra-modality learning aims to learn unimodal representations of tokens through their inherent attributes and contextual neighbors. In the image-text relevance measuring module, we use the global representations of both text sentence and vision image from previous stage to compute text-image matching score, and then have a local-global visual feature for the text based on the score. In the last, we iteratively perform cross-modal learning between vision and text to refine the most valuable feature for MNER. We conduct the extensive experiments and analyses to demonstrate the advantage of HamLearning.

\section{Acknowledgements}
This work was supported by the National Key Research and Development Program of China (2022YFB3103904), the National Natural Science Foundation of China (No.61931019), and the Strategic Priority Research Program of the Chinese Academy of Sciences (No. XDC02011200).

\bibliography{custom.bib}

\end{document}